\documentclass[a4paper,11pt]{article}

\usepackage{oloscribe}
\usepackage{dirtytalk}
\usepackage[ruled,vlined,linesnumbered]{algorithm2e}

\begin{document}

\makeheader{Pranshu Gupta}                     	      
           {Algorithms Inspired by Nature: A Survey}  						

\tableofcontents

\pagebreak
\section{Introduction}
Nature is known to be the best optimizer.  Natural processes 
most often than  not  reach  an  optimal  equilibrium.  Scientists  have always strived to  understand and model such processes. Thus, many algorithms exist today that are inspired by nature. Many of these algorithms and heuristics can be used to solve problems for which no polynomial time algorithms exist, such as Job Shop Scheduling and many other Combinatorial Optimization problems.  We  will  discuss  some  of these  algorithms  and  heuristics  and  how they  help  us  solve 
complex  problems  of  practical  importance.  We will focus mainly on Ant Colony  Optimization,  Particle  Swarm  Optimization,  Simulated Annealing  and  Evolutionary Algorithms.

\begin{remark}
A heuristic can be thought of as a rule of thumb that will hopefully find a good solution for a given problem. These techniques are mainly applied to complex convex multivariate constrained combinatorial optimization problems which are usually NP complete or harder.
\end{remark}

\begin{remark}
\say{A metaheuristic is a higher-level procedure or heuristic designed to find, generate, or select a heuristic that may provide a sufficiently good solution to an optimization problem, especially with incomplete or imperfect information or limited computation capacity.} [\cite{bianchi2009survey}]
\end{remark}

Ant Colony Optimization utilizes a family of heuristics inspired by the foraging behavior of Ants. Ants are highly co-operative social insects, they search for food in huge groups and communicate indirectly using a chemical called pheromone. These heuristics model individual ants as independent solution finders roaming in the solution space of a problem and communicating via a shared variable(s) i.e. the pheromone.

When a liquid is cooled slowly, its particles arrange themselves in some configuration that allows minimum energy state - this process is called annealing. Simulated Annealing tries to model this phenomenon by drawing an analogy between between the behavior of multi-body systems (e.g. liquids)  in thermal equilibrium at a finite temperature and the behaviour of the objective function of a combinatorial optimization problem at different values of the variables involved. [\cite{kirkpatrick1983optimization}]

Particle Swarm Optimization is a another optimization heuristic. It relies on a swarm of particles that are initialized with some configuration in the solution space of a given problem. These particles then explore the solution space individually while communicating with other particles around them to find the best solution. After some iterations these particles are expected to make swarms around optimal solution(s). PSO relies on randomization in the movement of particles to avoid getting stuck in local optima.

Evolutionary Algorithms are a family of algorithms inspired from Neo-Darwinism paradigm of biological evolution. 

\begin{equation*}
    \text{Neo-Darwininsm} = \text{Darwinism} + \text{Selectionism} + \text{Mendelian Genetics}
\end{equation*}

These algorithms model solutions as individual species with their genetic code representing the solution itself. These individuals later reproduce to create new individuals with different genetic code. Reproduction involves choosing healthy parents and then crossing over their genes to create an offspring. The optimality of the solution instance represented by the genetic code of the offspring determines its fitness. Then the notion of "survival of the fittest" comes into play. The best individuals survive and then they reproduce for the next generational cycle. Thus, this model utilizes both exploration and exploitation of the solution space. 

Genetic Algorithms, Evolutionary Strategies, Differential Evolution etc., are a few variants of EAs.  

\section{Ant Colony Optimization}
Ant Algorithms are a family of algorithms inspired by the behavior of real ant colonies. Ants are social insects and they always work together to ensure the survival of the colony as a whole. When in search of food, ants deposit a chemical substance called \textit{pheromone} on their path. If they succeed in finding a food source they again deposit pheromone on their path back to the colony. The amount of pheromone they deposit on the path often is determined by the quality and quantity of the food source that the path leads to. Also, if the multiple ants find the same food source, the one which followed the shorter path will deposit the pheromone first. When looking for food the ants also consider the pheromone deposited by other ants to aid their own search. Thus, the shorter paths and the paths which lead to better food sources will be followed by more and more ants and the pheromone on them will be further enriched (positive feedback in search). Also, because pheromone evaporates in atmosphere with the passage of time, paths which are no longer useful or followed will lose the pheromone deposited on them. However, ants are always free to choose their own paths ensuring the explore for new food sources as well instead of just clinging upon only one.

This behavior of ants can be utilized to design algorithms to solve search and optimization problems. \cite{dorigo1999ant} were the first to model foraging behavior of ant colonies and design a metaheuristic that can be used to solve problems of practical importance. 

\subsection{Ant Colony Optimization Metaheuristic}
The ACO metaheuristic is usually defined in the spirit of Combinatorial Optimization problems. It is indeed one of the family of approximate algorithms that are used to solve hard CO problems in reasonable amount of time. Let's define a CO problem.

\begin{definition}
A Combinatorial Optimization problem $P$ is defined as $P = (S, \omega, f)$ where $S$ is the solution space for the given problem, $\omega$ is the constraint set on the solution parameters and $f:S\rightarrow \bR^+$ is the objective function to be minimized. 
\end{definition}

The solution set can be defined as a set of feasible solutions. A feasible solution is a set of $N$ discrete variables $X_i \in D_i$ where $i \in {1,...n}$ which satisfy the solution constraints given by the problem instance. 

The pheromone trail model is the is the most crucial component of any ACO framework. It is used to probabilistically build the solution components of the problem instance. A solution component $c_{ij}$ is defined as the assignment of a variable $X_i$ to a value $v_j \in D_i$. The pheromone trail corresponding to this solution component is denoted by $T_{ij}$ with its value being $\tau_{ij}$. The set of all solution components is denoted by $C$ and the vector of pheromone trails is denoted by $T$.

For ACO, we define a complete graph which has a node corresponding to each possible solution component called the Construction Graph. In each iteration of ACO, each ant moves from one component to another effectively assigning values to the discrete variables (taking care of the constraints and considering deposited pheromone) and thus creating a feasible solution. At the end of every iteration, each ant would have constructed a feasible solution. Then, we compute the fitness of these solutions and deposit pheromone trail values on the edges that the ant which created the solution followed.

\hfill \break
\begin{algorithm}[H] 
\label{algorithm:aco}
\SetAlgoLined
    Initialize Pheromone Trail values $T$\\
    The best solution $S_{bs} \leftarrow$ NULL\\
    \While{termination conditions not met}{
        Set of solutions found in this iteration $S_{iter} \leftarrow \phi$\\
        \ForEach{ant in colony}{
            $s \leftarrow$ ConstructSolution($T$)\\
            \If{$s$ is feasible}{
                $s \leftarrow $ LocalSearch($s$) \\
                \If{$f(s) < f(s_{bs})$ OR $s_{bs}$ is NULL}{
                    $s_{bs} \leftarrow s$\\
                }
                $S_{iter} \leftarrow S_{iter} \cup \{s\}$\\
            }
        }
        ApplyPheromoneUpdate($T, S_{iter}, s_{bs}$) \\
    }
    Return $s_{bs}$ \\
    \caption{Generic ACO Algorithm [\cite{dorigo2005ant}]}
\end{algorithm}
\hfill \break

The probability that an ant will choose a solution component $c_{ij}$ as part of its solution conditioned on the set of solution components it has already chosen (the partial solution $s^p$) is given by:
\begin{equation*}
    p(c_{ij}|s^p) = \frac{\tau_{ij}^\alpha \cdot c_{ij}^\beta}{\sum_{c_{lk} \in N(s^p)} \tau_{lk}^\alpha \cdot c_{lk}^\beta } 
\end{equation*}
Here, $c_{ij}$ denotes the problem specific cost of choosing the assignment $j$ for decision variable $i$. $\alpha$ and $\beta$ are parameters than are used to tune the relative importance of the problem specific costs and pheromone trail value corresponding to solution component. $N(s^p)$ is the set of allowed solution components given that the partial solution $s^p$ has already been constructed. $N(s^p)$ depends on the constraint set of the problem instance. If the constraint set does not allow the selection of a solution component given the partial solution, the corresponding transition probability is 0.

The pheromone update rule is given as follows 
\begin{equation*}
    \tau_{ij} \leftarrow (1-\rho)\tau_{ij} + Q\sum_{k=1}^{m}\frac{1}{f(s_k)}
\end{equation*}
Here, $\rho$ is the evaporation coefficient which models the atmospheric evaporation of the pheromone, Q is some constant, $s_k$ is the solution constructed by the $k^{th}$ ant.

There are other variants of ACO which use different pheromone update rules. Consider the MIN-MAX ACO, the pheromone update rule for it is given by the following equation where $s_{iterbs}$ is the best solution found in this iteration. [\cite{dorigo2006artificial}]
\begin{equation*}
    \tau_{ij} \leftarrow (1-\rho)\tau_{ij} + \frac{1}{f(s_{iterbs})}
\end{equation*}
In the Ant Colony System Framework, instantaneous updates are also made to the pheromone trails. This slightly decreases the pheromone value allowing  other ants to possible find diverse solutions instead of premature convergence to a sub-optimal solution.
\begin{equation*}
    \tau_{ij} \leftarrow (1-\phi)\tau_{ij} + \phi\tau_0
\end{equation*}

\subsection{Applications of Ant Algorithms}
\subsubsection{Job Shop Scheduling}
A Job Shop Scheduling problem consists of a set of $n$ Jobs and a set of $m$ machines. Each job consists of a sequence operations each  to be processed on a specific machine. Now, because the resources have to be shared and each operation takes a given amount of time, we want to minimize the total time span required to finish all the jobs - the makespan. There are many techniques to solve this problem - Shifting Bottleneck, SA, ACO, PSO a well as EAs can be used to solve JSSP.

Every JSSP instance can be represented by a disjunctive graph [See Figure 1] in which we have one node for each operation with directed edges from $o_i$ to $o_j$ if the latter comes after the former in the constraint sequence given in the problem. These constraint sequences are generally given for operations that belong to the same job. We dont have a complete ordering across all the operations from different jobs. So, in the disjunctive graph we have un-directed edges between the operations belonging to different jobs but the same machine. If we define a machine specific ordering of operations on each machine, we have essentially solved the problem. We also have a start and finish node to denote the start and finish of the complete sequence of operations [\cite{yamada1997job}]

\begin{figure}
\hskip4ex
    \centering
	\includegraphics[width=0.6\textwidth]{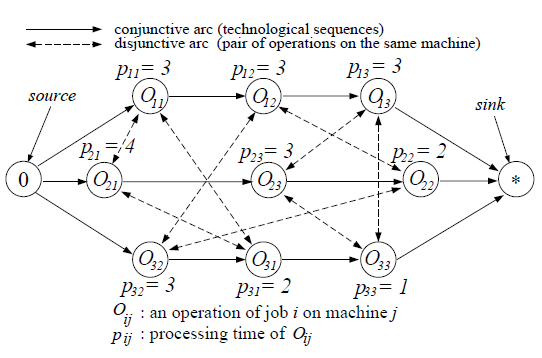} 
	\centering
	\caption{Disjunctive Graph for a JSSP with 3 machines and 3 Jobs each with 3 operations. $O_{ij}$ denotes an operation belonging to job $i$ which has to processed on machine $j$, $p_{ij}$ denotes the processing time corresponding to the operation. Image taken from [\cite{yamada1997job}]}
\hskip4ex
\end{figure}

In ACO, we create the construction graph for ant traversal using the disjunctive graph of the problem instance. We add two directed edges in the construction graph for each undirected edge in the disjunctive graph. An ant will choose on of those edges and hence the machine specific ordering will be found. Now, an ant walk from start node to the finish node in the construction graph will essentially give a complete solution. Note that the ant visits all the nodes in the graph one by one. The pheromone update rule captures the makespan as the fitness function and the transition probabilities respect the feasibility of the solution by making the probability zero if the transition creates a cycle in the ant walk, as a cyclic order is non-realizable. [\cite{blum2002ant}]

\subsubsection{Aircraft Conflict Resolution}
The Aircraft Conflict Resolution problem is of great practical importance. It is combinatorial optimization problem in which a set of $n$ aircrafts approaching a single point have to deviate from their original course in order to avoid collision.

\begin{figure}
\hskip4ex
    \centering
	\includegraphics[width=0.4\textwidth]{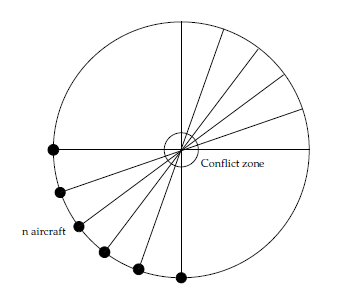} 
	\centering
	\caption{Aircraft Conflict Problem. Image taken from [\cite{durand2009ant}]}
\hskip4ex
\end{figure}

\cite{durand2009ant} have solved a simplified version of this problem in which all the aircrafts are approaching with the same speed, time is discretized, the set of possible manoeuvres is finte (say {$10^o, 20^o, 30^o$}) and only two manoeuvres are to be applied to each aircraft. Then this becomes a combinatorial optimization problem wherein we have to choose the manoeuvres at 2 time steps. The construction graph they have designed is shown in Figure 3. Each node is present at a discrete time step and there are three kinds of nodes U, V and W. The only allowed transition between differently typed nodes are U to V and V to W.

\begin{figure}
\hskip4ex
    \centering
	\includegraphics[width=0.7\textwidth]{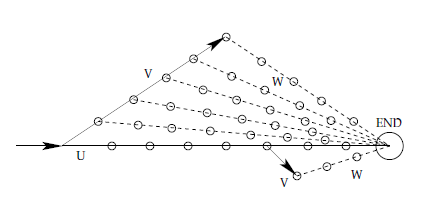} 
	\centering
	\caption{Construction Graph for Aircraft Conflict Problem. Image taken from [\cite{durand2009ant}]}
\hskip4ex
\end{figure}

Now, in each iteration of ACO, a set of $n$ ants create a single solution for the problem. Each of these ants corresponds to an aircraft and the walk done by the the ant defines the manoeuvres for the aircraft. We take this set on $n$ walks and see if they give a conflict free plan, if they do, the pheromone trail is updated according to the delay due to the course correction as the fitness function. They have demonstrated that this technique can resolve conflicts for upto 30 aircrafts in reasonable time.

\subsubsection{Ant Based Clustering}
Some species of ants are known to arrange their food supplies in clusters on the basis of different properties of the food items (size etc.). We can also model this behavior of ants to solve clustering problems as well.

\cite{handl2006ant} have proposed a simple ant based clustering algorithm that gives natural clusters. Unlike K-Means and other traditional clustering techniques which often need the programmer to specify the number of cluster centers beforehand, this technique gives natural clusters all by itself. They have experimentally shown that it is very robust and gives clusters that are inherent from the dataset, even if the clusters overlap.

\hfill \break
\begin{algorithm}[H] 
\label{algorithm:antcluster}
\SetAlgoLined
    Randomly scatter all the data points in a toroidal grid\\
    \ForEach{ant in colony}{
        Ant picks a randomly chosen datapoint ($i$)\\
        Ant places itself at a randomly selected location in the grid\\
    }    
    \While{termination conditions not met}{
        Choose a random ant and displace it by a given stepsize\\
        Drop the datapoint ($i$) at this location in the grid with probability given by
        \begin{equation*}
            p = \bigg(\frac{f^*(i)}{0.3 + f(i)}\bigg)^2
        \end{equation*}
        \If{datapoint was dropped}{
            \While{another datapoint is not picked}{
                Randomly choose a datapoint ($i$)\\
                Pick it up with the probability given by
                \begin{equation*}
                    p = \bigg(\frac{0.1}{0.3 + f(i)}\bigg)^2
                \end{equation*}                
            }
        }
    }
    \caption{Ant Based Clustering [\cite{handl2006ant}]}
\end{algorithm}
\hfill \break

The function $f(i)$ is the Lumer and Faieta neighborhood function defined as:
\begin{equation*}
    f(i) = max\bigg(0, \frac{1}{\sigma^2}\sum_{j \in L}\bigg(1 - \frac{\delta(i, j)}{\alpha}\bigg)\bigg)
\end{equation*}
where $\sigma^2$ is the size of local neighborhood in the toroidal grid, $\delta(i, j)$ is the distance function for the datapoints to be clustered, $L$ is the set of datapoints currently inside the $\sigma^2$ neighborhood in the grid. $\alpha$ is data specific scaling parameter. The modified LF function $f^*$ is given as follows:
\[
f^*(i) = 
    \begin{cases}
        f(i) &\quad\text{if $ \forall j$ $\big( 1 -  \frac{\delta(i, j)}{\alpha} \big) $}\\
        $0$ &\quad\text{otherwise}
    \end{cases}
\]
This modified LF neighborhood function inflicts heavy penalties for high dissimilarities thus improving the spatial separation between the clusters.

\pagebreak
\section{Simulated Annealing}
Annealing is the process of slowly cooling molten solids to get them in a crystallized state. If the object is cooled rapidly, its particle do not reach an equilibrium state and the crystals formed are small, thus reducing the overall strength of the object. Therefore, the cooling plan in this procedure is very important. In optimization settings, the simulated annealing algorithm we model the particles of the cooling object as the parameters of a possible solution of the given problem. The temperature is modeled as a variable which defines the transition probability of the selected solution. Each run of the algorithm consists of multiple iterations until an equilibrium (convergence) is reached. After that we change the transition probability of the solution by lowering the temperature. We stop the runs when the lower bound of temperature reached. See algorithm \ref{algorithm:simulateAnn}

\hfill \break
\begin{algorithm}[H] 
\label{algorithm:simulateAnn}
\SetAlgoLined
    Choose a random configuration $X_i$, select the initial system temperture $T_0$ and decide the cooling schedule (usually exponential)\\
    Evaluate Cost $C(X_i)$\\
    \While{temperature > lower bound}{
        Perturb $X_i$ to get a neighbor configuration $X_{i+1}$\\
        Evaluate Cost $C(X_{i+1})$\\
        \uIf{$C(X_{i+1}) < C(X_i)$}{
            Set the new solution configuration to $C(X_{i+1})$\\
        }
        \Else{
            Set the new solution configuration to $C(X_{i+1})$ with probability p
            \begin{equation*}
                p = e^{-(C(X_{i+1}) - C(X_i))/T_k}
            \end{equation*}
        }
        \If{Equilibrium reached at $T$}{
            Reduce the system temperature to 
            \begin{equation*}
                T_{k+1} = (\frac{T_1}{T_0})^k T_k
            \end{equation*}
        }        
    }
    \caption{Simulated Annealing}
\end{algorithm}

\subsection{Solving Job Shop Scheduling with Simulated Annealing}
The objective of the Job Shop Scheduling problem is to find out the optimal sequence of jobs on a given set of machines so that the makespan is minimized. JSSP is a NP Hard problem and Simulated Annealing can be used to solve it. \cite{van1992job} have shown that SA performs very well on JSSP instances. They modeled the configuration as the set of machine specific Job sequences with the makespan time as their costs. The neighborhood structure is defined as the set of configurations that can be obtained by reversing the order of two successive operations (on the same machine) on a critical path in the disjunctive graph. The perturbations in the SA give one of the neighbor solutions. This way the SA procedure can be adopted to solve any JSSP instance. Similarly, SA can be used to other combinatorial optimization problems like TSP and Physical Chip Design [See \cite{kirkpatrick1983optimization}]

\pagebreak
\section{Particle Swarm Optimization}
In Particle Swarm Optimization, a number of particles are initially scattered in the search space of a given problem. The fitness of each particle can be evaluated using its location in the search space. Now, in each iteration every particle displaces itself to a new position. This new position is determined by the history of best positions of this particle as well as some other particles in the swarm. It is in fact the weighted mean of the particle's own best position and the neighbors best position. The weights are picked uniformly from range of positive numbers. Eventually, the swarm as a whole is likely to move close to the optimum of the fitness function. See Algorithm \ref{algorithm:pso} for a generic PSO technique.

\hfill \break
\begin{algorithm}[H] 
\label{algorithm:pso}
\SetAlgoLined
    Initialize a population of particles at random positions in the search space\\
    \While{termination conditions not met}{
       For each particle $i$ evaluate the fitness function at its current position\\
       Update the best fitness value $pbest_i$ and position $p_i$ for this particle as per step 3\\
       Identify the best fitness particle $g$ in the neighborhood\\
       Update the position and velocity of the particle $i$ as per the following equations
       \begin{align*}
           v_i &\leftarrow v_i + U(0, \phi_1)\otimes(p_i - x_i) + U(0, \phi_2)\otimes(p_g - x_i)\\
           x_i &\leftarrow x_i + v_i
       \end{align*}
    }
    \caption{Particle Swarm Optimization}
\end{algorithm}
\hfill \break

This works quite well if the solution space of the problem is continuous, unlike the discrete valued solution spaces of problems like TSP, JSP etc. Defining velocities is a crucial part of the procedure, but according to \cite{clerc2010particle}, where a version of PSO for TSP is presented, velocity can be viewed as an operator which when applied to the current position gives a new position from the neighborhood.

There are many other variants of the Standard PSO, one in which the velocity of the particles is bounded, another in which the updates are determined not only by the particle's own history and its best neighbor but its entire neighborhood - Fully Informed Particle Swarm Optimization [\cite{poli2007particle}]. FIPSO has been demonstrated to work better than the standard PSO - but it is sensitive to the way the neighborhood is defined.

\subsection{Applications of Particle Swarm Optimization}
\subsubsection{Solving Job Shop Scheduling with Particle Swarm Optimization}
In PSO, the configuration and neighborhood definitions are the same as those in Simulated Annealing. However, we need to define the notion of velocity, the distance between two configurations and the sum of two velocities. The velocity itself can be defined as a permutation function that gives a new machine specific sequence of operations from the old one. Thus, the velocity can be defined as a list of pair of operations that belong to same machine, these pairs are to be exchanged in the machine specific ordering to get the new position i.e. a list of transposition. The distance between two positions is the velocity that must be applied to the first position to reach the second. Finally, the sum of two velocities is the order preserving truncated union of the transposition sets used to define the two velocities. 
With these definitions at hand we can go ahead and use PSO to solve JSSP. See \cite{shao2013hybrid}

\subsubsection{A Hybrid of SA and PSO to solve JSSP}
In a hybrid solution proposed by \cite{shao2013hybrid}, we initialize a large number of particles using an approach proposed by \cite{pezzella2008genetic}. The new position for each particle is found by simulated annealing approach. Thus, the structure of the overall algorithm remains like that of PSO (with SA embedded in it).

The advantage of this hybrid framework is that it takes the best from SA - local search and PSO - global search. Thus, avoiding local optima better than traditional SA, maintaining diversity in solutions and identifying better neighbors than the traditional PSO.

\section{Evolutionary Algorithms}
Evolutionary Algorithms are a class of general purpose algorithms that are designed to solve optimization problems that are large, complex, discontinuous, non-differentiable, multi-modal and possibly constrained as well. EAs are highly parallelizable, simple and are known to almost always reach a near optimum or the global optimum.

Any Evolutionary Algorithm generally comprises of the following components - a population generator, a population selector, fitness estimator and reproduction operators (crossover and mutation). The pseudocode of a generic EA can be given as Algorithm \ref{algorithm:genEA}
\hfill \break

\begin{algorithm}[H] 
\label{algorithm:genEA}
\SetAlgoLined
    $t=0$\\
    Randomly generate initial population $P(0)$\\
    \While{termination condition not met}{
        Evaluate fitness of each individual of $P(t)$\\
        Select individuals as parents from $P(t)$ based on their fitness\\
        Apply search operators (crossover and mutation) on parents and generate $P(t+1)$\\
        $t:=t+1$\\
    }
    \caption{A Generic Evolutionary Algorithm [\cite{du2016search}]}
\end{algorithm}

\hfill \break
$P(t)$ denotes the current set of individuals in the population. Each individual in the population is called a chromosome which represents a solution to the given problem. The chromosome is denoted as a string of elements called genes which represent individual parameters of the solution. An individual's chromosome is termed as the genotype and genotype when denoted along with fitness and other properties of the corresponding solution is called a phenotype.

\subsection{Encoding a solution as a Chromosome}
There are different techniques to represent a solution as a chromosome, each having their pros and cons. The performance of these algorithms depends on the encoding technique used, because the crossover and mutation operators are applied on these representations only to explore the solution space. Some of the notable representation techniques are Binary Coding, Gray Coding, Delta Coding and Fuzzy Coding.
\subsubsection{Binary Coding}
Genetic Algorithms use Binary Encoding wherein each chromosome is an array of genes $x$. Each of these genes $x_i$ represents some parameter of the solution in form of binary string of length $l$. To recover the actual value of the parameter whose value lies in the interval $[x_i^{-}, x_i^{+}]$ we use Equation \ref{equation:binCode} [\cite{du2016search}]
\begin{equation} \label{equation:binCode}
    x_i = x_i^{-} + (x_i^{+} - x_i^{-}) \frac{1}{2^{l_i}-1}(\sum_{j=0}^{l_i-1}s_j 2^i)
\end{equation} 
Binary Coding is a fixed length and static coding which sometimes leads to slower convergence and lower accuracy when it reaches near the optimal solution. Therefore, new dynamic coding techniques were developed, such as Delta and Fuzzy Coding.

\subsubsection{Delta Coding}
Delta Coding is a dynamic coding technique in which the representation of the search space changes in after each iteration. It starts with Binary Coding but modifies the representation by changing the bit length of the solution parameters after each iteration. The solution parameters in subsequent iterations are denoted with respect to the $S_{best}$ from the previous iteration. The distance from $S_{best}$ is called the Delta Value. This creates a new hypercube with $S_{best}$ as its origin at each iteration which is then searched using the crossover operators. In order to maintain diversity in the population in order to explore the search space, we re-initialize the population in each iteration. See Algorithm \ref{algorithm:delta} [\cite{Mathias:1994:CRD:1326668.1326672}].
\hfill \break

\begin{algorithm}[H] 
\label{algorithm:delta}
\SetAlgoLined
    Initialize the population randomly $P(0)$\\
    \While{trials $<$ max.Trials $\&$ fitness $>$ threshold}{
        \While{Diversity Metric $>$ 1}{
            Apply Crossover Operators to get candidates for $P(t+1)$\\
            Evaluate fitness function for each individual\\
            Insert valid off-springs to $P(t+1)$\\
        }
        Set the best new solution as $S_{best}$\\
        Re-initialize Population\\
        \uIf{Delta Value $== 0$}{
            \If{Parameter Length $>$ Original Parameter Length}{
                Encode parameter using $1$ additional bit
            }    
        }\Else{
            \If{Parameter Length $>$ Lower Bound}{
                Encode parameter using $1$ less bit
            }               
        }
    }
    \caption{Delta Coding}
\end{algorithm}

\hfill \break
\subsubsection{Fuzzy Coding}
In fuzzy coding, for each parameter of the solution we choose one or more from a set of fuzzy sets (NM, Ms, ZR, PS, PM). Each solution parameter has a membership function (fuzzy sigmoid, fuzzy gaussian or fuzzy normal). These membership functions are used to choose values from selected fuzzy sets corresponding to the solution parameter. The value of the solution parameter is denoted by the weighted mean of these chosen vales as per the membership function weights. The membership functions for the fuzzy set NM are given as follows [\cite{sharma2003fuzzy}]:
\begin{align*} 
    \text{Fuzzy Sigmoid}: y &= -ae^{-(\frac{x-c}{\frac{d}{3}})^2} - b - 1 \\ 
    \text{Fuzzy Gaussian}: y &= ae^{-(\frac{x-c}{\frac{d}{3}})^2} + b\\
    \text{Fuzzy Normal}: y &= 1 - \frac{x-c}{d}
\end{align*}
IN this coding technique, the genetic algorithm need only optimize the fuzzy sets and the membership function corresponding to each solution parameter. As the actual value of the parameter is indirectly obtained using the chosen fuzzy sets and the membership function, the genetic operators (crossover and mutation) are independent of the actual values. We only need to specify the type of the parameter and its range beforehand.

\subsection{Selection and Replacement}
In every iteration, we need to choose the set of parents from the population, this can be done by using various policies. Some of the important selection policies are Roulette Wheel Selection, Ranking Selection, Tournament Selection and Elitist Selection. 
\subsubsection{Roulette Wheel Selection}
In this policy, the selection probability of an individual is proportional to its fitness value. We can think of this policy as a roulette wheel which is partitioned as per the fitness of each individual.
\begin{align*}
P_i &= \frac{f(x_i)}{\sum_{i=1}^{N_P}f(x_i)} & i &= 1,2,..., N_P    
\end{align*}
\subsubsection{Ranking Selection}
In this policy, the individuals are sorted in ascending order according to their fitness and the best one is assigned rank 1. This policy applies a uniform selective pressure on all the individuals unlike the roulette wheel policy. The probability of the selection of $i^{th}$ individual as a parent is given by:
\begin{align*}
    P_i &= \frac{1}{N_P}(\beta - 2(\beta - 1)\frac{i-1}{N_P - 1}) & \beta &\in [0,2]
\end{align*}
\subsubsection{Tournament Selection}
In this policy two individuals $i$ and $j$ are picked up randomly with replacement. The probability that the individual $i$ is selected is given by:
\begin{align*}
    P_i = \frac{1}{1+exp(\frac{f_j - f_i}{T})}
\end{align*}
We stop after the required number of parents have been selected.
\subsubsection{Replacement}
After choosing the parents and obtaining the offspring, we replace some individuals from the population. There are several policies to accomplish this too - complete generational replacement, random replacement, replace worst policy, replace oldest policy and replce the similar parent policy.

\subsection{Reproduction}
Reproduction is simulated by Crossover and Mutation operators. Crossover operator takes two individuals as parents and gives two new individuals by swapping some parts of the chromosomes of the parents. Mutation is a unary operator,it needs only one individual as a parent to create an offspring.

There are different types of crossover operators - one point crossover, two point crossover, multi-point crossover and uniform crossover. [See Figure 4]
Uniform Crossover swaps are more disruptive and exploratory hence, suitable for small populations. Two point crossover is generally good for large populations, It has been demonstrated in many experiments that Two Point crossover works better than One Point crossover in most of the cases.

Mutation is another genetic operator it randomly chooses one or more bits from the chromosome and flips them to get a new individual. Mutations introduce the necessary noise to perform hill climbing in genetic algorithms. 

\begin{figure}
\hskip4ex
    \centering
	\includegraphics[width=0.5\textwidth]{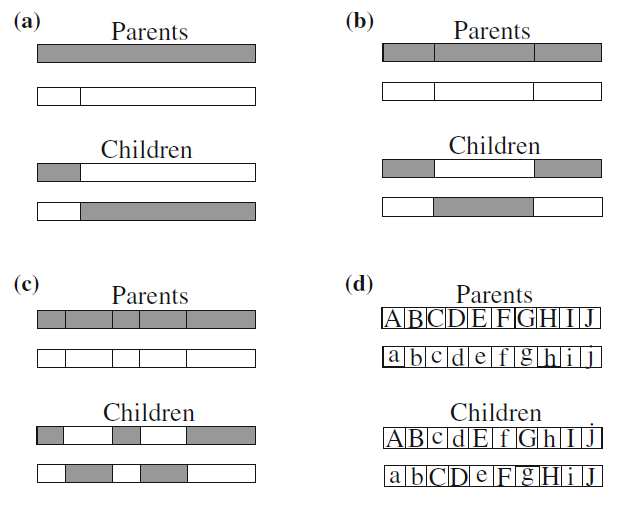} 
	\centering
	\caption{Crossover Operators - a) One Point b) Two Point c) Multi Point d) Uniform Crossover. Image taken from [\cite{du2016search}]}
\hskip4ex
\end{figure}

\subsection{Applications of Evolutionary Algorithms}
Evolutionary Algorithms have many applications in search and optimization. Some variants of EAs are clubbed with machine learning algorithms to find their optimal tuning parameters so as to find the best model. EAs can perform symbolic regression which are often very difficult for traditional regression algorithms.
\subsubsection{Symbolic Regression with Genetic Algorithms}
Genetic Programming is a variant of Evolutionary Algorithm which can be used for symbolic regression. The encoding used in genetic programming is in form of hierarchical trees. We can represent mathematical function with the help of these trees. See Figure 5. Now, all weed to perform symbolic regression is to define the crossover and mutation operators. See Figure 6. 
\begin{figure}
\hskip4ex
    \centering
	\includegraphics[width=0.2\textwidth]{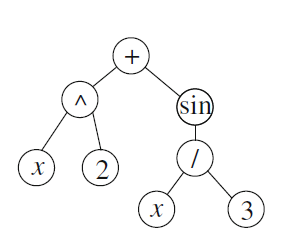} 
	\centering
	\caption{Tree Encoding for $f(x) = x^2 + sin(x/3)$. Image taken from  [\cite{du2016search}]}
\hskip4ex
\end{figure}

\begin{figure}
\hskip4ex
    \centering
	\includegraphics[width=0.5\textwidth]{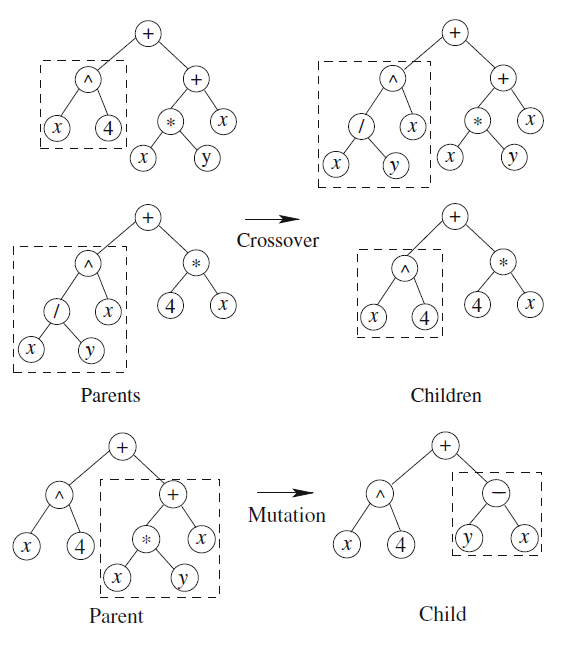} 
	\centering
	\caption{Crossover and Mutation in Symbolic Regressio. Image taken from [\cite{du2016search}]}
\hskip4ex
\end{figure}

\subsubsection{Support Vector Regression with Genetic Algorithms}
\cite{chen2007support} have used Genetic Algorithms for tuning the parameters of the traditional Support Vector Regression algorithm to forecast tourism data. Neural Networks have been shown to work very well on this problem. However, they have some drawbacks such as large number of parameters to train, risk of overfitting and difficulty in obtaining a stable solution. Support Vector Regression is also known to work excellently for non-linear regression problems but it needs careful setting of its parameters to avoid bad performance and overfitting. But we don't have any general rule to tune these parameters, it comes down to knowledge of the experimenter and trial and error methods for this. 

This is the part where genetic algorithms come into picture, \cite{chen2007support} applied GA to find the optimal parameters for SVR. Their results are at par with the Neural Network framework used for the tourism forecasting. 

\section{Conclusion}
We have discussed some very beautiful algorithms inspired by nature which can be used to solve search and optimization problems. Also, the problems we discussed are of great practical importance, take Aircraft Conflict Resolution as an example and Job Shop Scheduling. Solving these problems can save lives on one hand and time and money on the other. Yet, this is not all, there is a whole world of such algorithms, algorithms which were designed by observing and modelling the miracles that nature does. Algorithms like Bee Hive Optimization, Cuckoo Search, Bacterial Foraging, Glow-worm Swarm Optimization, Biomolecular Computing and so on, the list is almost endless. All of them helping us to solve complex problems, mostly problems which are NP Hard.

\bibliographystyle{plainnat}
\bibliography{references}

\end{document}